\documentclass{article}
\usepackage[utf8]{inputenc}
\usepackage{amsmath,amssymb,amsfonts}
\usepackage{graphicx}
\usepackage{booktabs}
\usepackage{hyperref}
\usepackage{natbib}
\usepackage[margin=1in]{geometry}
\usepackage{subcaption}
\usepackage{xcolor}

\title{Leap+Verify: Regime-Adaptive Speculative Weight Prediction\\for Accelerating Neural Network Training}

\author{Jeremy McEntire}

\date{}

\begin{document}
\maketitle

\begin{abstract}
We introduce \textbf{Leap+Verify}, a framework that applies speculative execution---predicting future model weights and validating predictions before acceptance---to accelerate neural network training.
Inspired by speculative decoding in language model inference and by the Automatically Scalable Computation (ASC) architecture for program execution, Leap+Verify decomposes training into three dynamically detected regimes (chaotic, transition, stable) using activation-space cosine similarity as a real-time Lyapunov proxy signal.
Within each regime, analytic weight predictors (momentum, linear, quadratic extrapolation) attempt to forecast model parameters $K$ training steps ahead; predictions are accepted only when validated against a held-out loss criterion.

We evaluate Leap+Verify on GPT-2 124M and Qwen 2.5-1.5B trained on WikiText-103 across five random seeds, sweeping prediction depth $K \in \{5, 10, 25, 50, 75, 100\}$.
Momentum-based prediction (Adam moment extrapolation) fails catastrophically at \emph{both} scales, with predicted losses exceeding actuals by $100{-}10{,}000\times$---a universal \textbf{norm explosion} in optimizer-state extrapolation.
Finite-difference predictors (linear, quadratic) succeed where momentum fails: at 124M, they achieve 24\% strict acceptance at $K{=}5$ in stable regimes; at 1.5B, they achieve 37\% strict acceptance in transition regimes.
The \textbf{scale-dependent} finding is in regime distribution: GPT-2 124M spends 34\% of training in stable regime, while Qwen 1.5B spends 64\% in chaotic regime and reaches stable in only 0--2 of 40 checkpoints.
Larger models are \emph{more predictable when predictable}, but \emph{less often predictable}---the practical bottleneck shifts from predictor accuracy to regime availability.
Cross-seed results are highly consistent ($<$1\% validation loss variance), and the three-regime framework produces identical phase boundaries ($\pm 50$ steps) across seeds.
\end{abstract}

% ============================================================
\section{Introduction}
\label{sec:intro}
% ============================================================

Training modern neural networks is an inherently sequential process: each gradient update depends on the current parameter state, which depends on all prior updates.
This sequential bottleneck consumes enormous computational resources---training frontier language models requires thousands of GPU-hours and millions of dollars~\citep{kaplan2020scaling}.
Yet the sequential nature of gradient descent is not as rigid as it appears.
Weight trajectories exhibit substantial regularity, particularly in late training, suggesting that future parameter states may be predictable from past trajectory information.

\textbf{Speculative execution} offers a principled way to exploit such predictability.
In speculative decoding for language model inference~\citep{leviathan2023fast,chen2023accelerating}, a small draft model generates candidate tokens that the target model verifies in parallel, achieving 2--3$\times$ speedup without altering the output distribution.
The key insight is the \emph{verify-then-accept} mechanism: predictions are cheap to generate and cheap to validate, so even moderate prediction accuracy yields substantial speedup.

We transplant this mechanism from inference to training.
\textbf{Leap+Verify} predicts future model weights $K$ steps ahead using analytic extrapolation of the weight trajectory, then validates the predicted weights against a held-out loss criterion before accepting the advance.
The name derives from the Automatically Scalable Computation (ASC) architecture~\citep{waterland2014asc}, which accelerates sequential program execution by speculatively predicting future program states and caching verified state transitions.
In ASC, a \emph{recognizer} identifies predictable states, \emph{predictors} forecast future states, and a \emph{validator} confirms correctness before fast-forwarding execution.
Leap+Verify instantiates each component for neural network training: the recognizer is an activation-space regime detector, the predictors are analytic weight extrapolators, and the validator is a held-out loss comparison.

A central finding is that prediction viability depends critically on the \emph{training regime}.
We identify three regimes---chaotic, transition, and stable---using cosine similarity between consecutive activation fingerprints as a proxy for the local Lyapunov exponent.
Prediction is viable only in transition and stable regimes, where weight trajectories exhibit sufficient regularity.
This regime dependence motivates \emph{conditional} prediction: the system predicts only when the regime detector indicates favorable conditions, avoiding wasted computation during chaotic training phases.

Our most striking empirical finding is a \textbf{universal momentum catastrophe}: Adam moment extrapolation produces weight predictions $100{-}10{,}000\times$ worse than actual validation loss at both 124M and 1.5B parameters.
This catastrophe is caused by norm explosion---the extrapolated displacement $K \cdot m_t / \sqrt{v_t}$ far exceeds the region of validity around the current point on the loss surface.
Finite-difference predictors (linear, quadratic), which extrapolate from actually observed checkpoint deltas, avoid this failure and achieve 9--37\% strict acceptance at $K{=}5$ depending on regime and scale, with proximity-based acceptance reaching 100\% at short horizons.
The \textbf{scale-dependent} finding is that the distribution of training regimes shifts dramatically: at 1.5B, training remains chaotic for 64\% of checkpoints (vs.\ 4\% at 124M), severely limiting when prediction is viable.
This regime-availability bottleneck has not been identified in prior work on weight nowcasting~\citep{jang2023wnn,knyazev2025nino,guan2024xgrad}.

\paragraph{Contributions.}
\begin{enumerate}
    \item \textbf{Leap+Verify}: A verify-then-accept mechanism for speculative weight prediction during training, inspired by speculative decoding and ASC.
    \item \textbf{Regime-conditional prediction}: Three dynamically detected training regimes (chaotic, transition, stable) that determine when and how to predict.
    \item \textbf{Universal momentum catastrophe and scale-dependent regime distribution}: Empirical demonstration that optimizer-state extrapolation fails at all scales, while the regime availability bottleneck grows with model size.
    \item \textbf{Reproducible evaluation}: Five-seed experiments with consistent results across GPT-2 124M and Qwen 2.5-1.5B on WikiText-103.
\end{enumerate}

% ============================================================
\section{Background and Motivation}
\label{sec:background}
% ============================================================

\subsection{Automatically Scalable Computation}

The Automatically Scalable Computation (ASC) architecture~\citep{waterland2014asc,waterland2013computational} accelerates sequential program execution by viewing it as a trajectory through state space.
The architecture has three core components: (1) a \emph{recognizer} that identifies states amenable to prediction, (2) a set of \emph{predictors} that learn to forecast future states from observed trajectory structure, and (3) a \emph{trajectory cache} that stores verified start-state/end-state pairs for fast-forwarding.
When the recognizer identifies a predictable state, predictors generate candidate future states, speculative threads execute from those candidates, and results are cached.
The sequential execution thread periodically queries the cache; on a match, it fast-forwards to the cached end state, achieving speedup proportional to prediction accuracy and trajectory length~\citep{waterland2014asc}.

ASC was originally implemented for x86 binary programs, where state vectors represent processor registers and memory.
The present work adapts the architecture for neural network training, where the ``state vector'' is the model's parameter tensor and the ``transition function'' is a gradient update step.
The exponentially large state space of parameters becomes tractable because training trajectories, like program trajectories, exhibit regular structure that predictors can exploit.

\subsection{Speculative Decoding}

Speculative decoding~\citep{leviathan2023fast,chen2023accelerating,stern2018blockwise} accelerates autoregressive language model inference by using a small draft model to generate candidate token sequences that the target model verifies in parallel.
The critical property is that verification is cheaper than generation: checking whether a sequence of $K$ tokens is acceptable requires a single forward pass, while generating them sequentially requires $K$ passes.
Acceptance is governed by a rejection sampling scheme that preserves the target distribution exactly.

Leap+Verify adapts the predict-then-verify template to training.
The ``draft model'' is an analytic weight extrapolator (momentum, linear, or quadratic); the ``verification'' is a loss evaluation on held-out data.
Unlike speculative decoding, where acceptance is binary and distribution-preserving, Leap+Verify uses graded acceptance criteria (strict, adaptive, proximity-based) that trade off between conservatism and skip distance.

\subsection{Training Dynamics and Phase Transitions}

Neural network training traverses qualitatively different regimes.
\citet{cohen2021gradient} identified two phases in full-batch gradient descent: progressive sharpening (where the maximum Hessian eigenvalue rises to $2/\eta$) and edge-of-stability (where sharpness oscillates at this threshold).
\citet{lewkowycz2020large} documented a ``catapult phase'' where loss temporarily grows before decreasing.
\citet{frankle2020linear} showed that networks become ``stable to SGD noise'' early in training, after which models sharing a common training prefix converge to linearly connected solutions.

These observations suggest a multi-phase training structure, but prior work detects phases using expensive Hessian computations~\citep{cohen2021gradient} or post-hoc weight-space analysis~\citep{frankle2020linear}.
We use activation-space cosine similarity as an efficient, real-time proxy that requires only forward passes on a fixed probe set.

\subsection{Weight Nowcasting}

Several lines of work predict future weights during training.
\citet{kamarthi1999accelerating} used Taylor series extrapolation of weight trajectories.
\citet{sinha2017introspection} trained a neural network to forecast future weights from weight history.
\citet{jang2023wnn} designed the Weight Nowcaster Network (WNN), predicting 5 epochs ahead from 5 epochs of history.
\citet{knyazev2025nino} improved this with graph neural networks modeling neuron connectivity (NiNo), achieving up to 50\% training acceleration.
\citet{guan2024xgrad} proposed XGrad, predicting future weights using optimizer update rules.

All existing weight prediction methods apply predictions \emph{unconditionally}---they do not verify predictions before acceptance, and they do not condition on detected training regimes.
Leap+Verify introduces both mechanisms: verify-then-accept prevents bad predictions from corrupting training, and regime conditioning restricts prediction to phases where trajectories are sufficiently regular.

% ============================================================
\section{Method}
\label{sec:method}
% ============================================================

\subsection{Regime Detection via Activation Fingerprinting}
\label{sec:regime}

We detect training regimes by measuring the stability of the model's internal representations.
At each checkpoint (every 50 training steps), we compute an \emph{activation fingerprint}: the concatenated final hidden states produced by a fixed set of 100 probe sentences.
The cosine similarity between consecutive fingerprints serves as a proxy for the local Lyapunov exponent of the training trajectory.

Formally, let $\mathbf{a}_t$ denote the activation fingerprint at step $t$.
We compute:
\begin{equation}
    s_t = \frac{\mathbf{a}_t \cdot \mathbf{a}_{t - \Delta}}{\|\mathbf{a}_t\| \|\mathbf{a}_{t-\Delta}\|}
\end{equation}
where $\Delta = 50$ (the checkpoint interval).
We classify regimes using thresholds derived from initial training runs:
\begin{equation}
    \text{regime}(t) = \begin{cases}
        \text{stable} & \text{if } s_t > \tau_{\text{high}} \\
        \text{chaotic} & \text{if } s_t < \tau_{\text{low}} \\
        \text{transition} & \text{otherwise}
    \end{cases}
\end{equation}
where $\tau_{\text{high}}$ and $\tau_{\text{low}}$ are averaged across initial seeds.

The cosine similarity signal offers several advantages over alternatives.
Unlike Hessian eigenvalue computation~\citep{cohen2021gradient}, it requires only forward passes (no second-order gradients).
Unlike weight-space linear interpolation~\citep{frankle2020linear}, it operates in the representation space where functional similarity is more directly measured.
The probe set is fixed across all checkpoints, ensuring that changes in $s_t$ reflect changes in the model's representations rather than changes in input distribution.

\subsection{Speculative Weight Prediction}
\label{sec:prediction}

Given a checkpoint at step $t$ with parameters $\theta_t$, we predict parameters $\theta_{t+K}$ using three analytic predictors that exploit different trajectory properties:

\paragraph{Momentum prediction.}
Uses the exponential moving averages maintained by the Adam optimizer:
\begin{equation}
    \hat{\theta}_{t+K}^{\text{mom}} = \theta_t + K \cdot \frac{m_t}{\sqrt{v_t} + \epsilon}
\end{equation}
where $m_t$ and $v_t$ are Adam's first and second moment estimates.
This extrapolates the current update direction at constant velocity.

\paragraph{Linear prediction.}
Fits a line to two consecutive checkpoint states:
\begin{equation}
    \hat{\theta}_{t+K}^{\text{lin}} = \theta_t + \frac{K}{\Delta}(\theta_t - \theta_{t-\Delta})
\end{equation}
This extrapolates the finite-difference velocity of the parameter trajectory.

\paragraph{Quadratic prediction.}
Fits a parabola through three consecutive checkpoints:
\begin{equation}
    \hat{\theta}_{t+K}^{\text{quad}} = \theta_t + \frac{K}{\Delta}(\theta_t - \theta_{t-\Delta}) + \frac{K(K-\Delta)}{2\Delta^2}(\theta_t - 2\theta_{t-\Delta} + \theta_{t-2\Delta})
\end{equation}
This captures curvature in the trajectory, accounting for acceleration/deceleration.

After computing $\hat{\theta}_{t+K}$, we load the predicted weights into the model and evaluate the validation loss $\hat{L}_{t+K}$.
We compare this against the current validation loss $L_t$ using three acceptance criteria:
\begin{itemize}
    \item \textbf{Strict}: Accept if $\hat{L}_{t+K} < L_t$ (predicted improvement).
    \item \textbf{Adaptive}: Accept if $\hat{L}_{t+K} < L_t + \sigma_L$ (within one standard deviation of recent validation losses).
    \item \textbf{Proximity (pct)}: Accept if $|\hat{L}_{t+K} - L_t| < \epsilon \cdot L_t$ (within a percentage of current loss).
\end{itemize}

If accepted, the model fast-forwards to step $t+K$, skipping $K$ gradient updates.
If rejected, training continues from step $t$ with no modification---the prediction is purely speculative with no side effects.

\subsection{Cascaded Prediction}
\label{sec:cascade}

To test deeper speculation, we chain multiple predictions in sequence.
A cascade of depth $D$ with step size $K$ applies prediction $D$ times, potentially advancing $D \times K$ steps.
Each stage uses the predicted weights from the previous stage as its starting point.
Cascades are evaluated only from stable-regime checkpoints, where individual prediction accuracy is highest.

% ============================================================
\section{Experimental Setup}
\label{sec:experiments}
% ============================================================

\subsection{Models and Data}

We evaluate on two language models of different scale:
\begin{itemize}
    \item \textbf{GPT-2 124M}: 12 layers, 768 hidden, 12 heads. Randomly initialized, trained from scratch.
    \item \textbf{Qwen 2.5-1.5B}: 28 layers, 1536 hidden, 12 heads. Randomly initialized from pretrained architecture config, trained from scratch.
\end{itemize}
Both models are trained on WikiText-103~\citep{merity2017pointer} with sequence length 256, using AdamW ($\eta = 5 \times 10^{-5}$, $\beta_1 = 0.9$, $\beta_2 = 0.999$, weight decay 0.01) with cosine learning rate schedule and 100-step warmup, for 2000 steps.

\subsection{Evaluation Protocol}

For each model, we run five seeds (42--46) with identical hyperparameters.
Each run proceeds in three passes:
\begin{enumerate}
    \item \textbf{Pass 1 (Training)}: Train for 2000 steps, saving checkpoints every 50 steps (40 checkpoints per seed). Record activation fingerprints and regime classifications.
    \item \textbf{Pass 2 (K-Sweep)}: For each non-chaotic checkpoint, evaluate all three predictors at $K \in \{5, 10, 25, 50, 75, 100\}$ with all three acceptance criteria. This yields up to $40 \times 3 \times 6 \times 3 = 2160$ evaluations per seed.
    \item \textbf{Pass 3 (Cascades)}: From stable-regime checkpoints, evaluate cascaded predictions with configurations $(D, K) \in \{(4, 25), (2, 50), (10, 10)\}$.
\end{enumerate}

% ============================================================
\section{Results}
\label{sec:results}
% ============================================================

\subsection{Training and Regime Detection}
\label{sec:results:regimes}

Table~\ref{tab:regimes} summarizes the regime breakdown across both model scales.
Both models were trained for 2000 steps with checkpoints every 50 steps (40 checkpoints per seed).
Regime classification is based on activation-space cosine similarity (Section~\ref{sec:regime}).

\begin{table}[h]
\centering
\caption{Regime breakdown by model scale (mean $\pm$ std across 5 seeds).
}
\label{tab:regimes}
\begin{tabular}{lcccc}
\toprule
\textbf{Model} & \textbf{Chaotic} & \textbf{Transition} & \textbf{Stable} & \textbf{Unknown} \\
\midrule
GPT-2 124M     & $1.6 \pm 1.7$ & $24.0 \pm 2.7$ & $13.4 \pm 3.2$ & $1.0 \pm 0.0$ \\
Qwen 2.5-1.5B  & $25.6 \pm 0.5$ & $12.4 \pm 1.1$ & $1.0 \pm 1.0$ & $1.0 \pm 0.0$ \\
\bottomrule
\end{tabular}
\end{table}

The regime distribution shifts dramatically with model scale.
GPT-2 124M spends 34\% of training in stable regime and only 4\% in chaotic, while Qwen 2.5-1.5B spends 64\% in chaotic and reaches stable in only 0--2 of 40 checkpoints.
The first unknown checkpoint in each run reflects the initial state before consecutive fingerprints are available for similarity computation.
Cross-seed regime boundaries are highly consistent at both scales: the chaotic-to-transition boundary occurs within $\pm 50$ steps across all seeds.

GPT-2 124M achieves a final validation loss of $1.616 \pm 0.007$ across seeds.
Qwen 2.5-1.5B achieves $0.810 \pm 0.019$ (range: 0.797--0.843), with training times of $34.9 \pm 0.7$ minutes per seed on A100 40GB with mixed precision.

\subsection{K-Sweep: Speculative Depth vs.\ Acceptance Rate}
\label{sec:results:ksweep}

Tables~\ref{tab:strict} and~\ref{tab:pct} present acceptance rates across prediction depth $K$, predictor type, and training regime.
Evaluations are performed only on non-chaotic checkpoints (transition + stable).

\begin{table}[h]
\centering
\caption{Strict acceptance rate (\%) in transition regime: predicted loss must improve over current loss.
Values are mean $\pm$ std across 5 seeds.
$N$ = total checkpoint evaluations across seeds.
}
\label{tab:strict}
\begin{tabular}{r|ccc}
\toprule
$K$ & Momentum & Linear & Quadratic \\
\midrule
\multicolumn{4}{c}{\textit{GPT-2 124M --- Transition regime}} \\
\midrule
5   & $0.0 \pm 0.0$ & $9.3 \pm 3.8$  & $7.9 \pm 3.0$  \\
10  & $0.0 \pm 0.0$ & $3.3 \pm 3.2$  & $3.4 \pm 3.3$  \\
25  & $0.0 \pm 0.0$ & $0.7 \pm 1.7$  & $0.0 \pm 0.0$  \\
\midrule
\multicolumn{4}{c}{\textit{GPT-2 124M --- Stable regime}} \\
\midrule
5   & $10.0 \pm 8.4$ & $24.3 \pm 6.8$ & $22.1 \pm 10.8$ \\
10  & $10.0 \pm 8.4$ & $18.8 \pm 8.0$ & $13.9 \pm 8.4$  \\
25  & $10.0 \pm 8.4$ & $5.8 \pm 3.5$  & $9.1 \pm 2.0$   \\
\midrule
\multicolumn{4}{c}{\textit{Qwen 2.5-1.5B --- Transition regime}} \\
\midrule
5   & $0.0 \pm 0.0$ & $37.2 \pm 11.0$ & $36.5 \pm 12.5$ \\
10  & $0.0 \pm 0.0$ & $31.3 \pm 4.7$  & $39.4 \pm 9.1$  \\
25  & $0.0 \pm 0.0$ & $22.9 \pm 8.1$  & $20.5 \pm 9.2$  \\
50  & $0.0 \pm 0.0$ & $17.8 \pm 9.0$  & $13.2 \pm 6.3$  \\
75  & $0.0 \pm 0.0$ & $7.9 \pm 8.4$   & $6.7 \pm 9.9$   \\
100 & $0.0 \pm 0.0$ & $6.2 \pm 9.1$   & $2.2 \pm 5.0$   \\
\bottomrule
\end{tabular}
\end{table}

\begin{table}[h]
\centering
\caption{Proximity (pct) acceptance rate (\%) in transition regime: predicted loss within 5\% of current loss.
Values are mean $\pm$ std across 5 seeds.
}
\label{tab:pct}
\begin{tabular}{r|ccc}
\toprule
$K$ & Momentum & Linear & Quadratic \\
\midrule
\multicolumn{4}{c}{\textit{GPT-2 124M --- Transition regime}} \\
\midrule
5   & $0.0 \pm 0.0$ & $99.1 \pm 1.9$ & $100.0 \pm 0.0$  \\
10  & $0.0 \pm 0.0$ & $97.3 \pm 2.5$  & $93.4 \pm 4.2$   \\
25  & $0.0 \pm 0.0$ & $44.0 \pm 9.8$  & $37.8 \pm 12.0$  \\
50  & $0.0 \pm 0.0$ & $14.1 \pm 6.8$  & $9.3 \pm 7.1$    \\
\midrule
\multicolumn{4}{c}{\textit{GPT-2 124M --- Stable regime}} \\
\midrule
5   & $16.5 \pm 4.6$ & $98.9 \pm 2.5$  & $98.9 \pm 2.5$  \\
10  & $16.5 \pm 4.6$ & $95.3 \pm 4.7$  & $97.6 \pm 3.4$  \\
25  & $16.5 \pm 4.6$ & $74.6 \pm 2.6$  & $70.6 \pm 7.7$  \\
50  & $14.8 \pm 5.3$ & $54.1 \pm 16.0$ & $48.2 \pm 20.2$ \\
\midrule
\multicolumn{4}{c}{\textit{Qwen 2.5-1.5B --- Transition regime}} \\
\midrule
5   & $0.0 \pm 0.0$ & $100.0 \pm 0.0$  & $100.0 \pm 0.0$  \\
10  & $0.0 \pm 0.0$ & $100.0 \pm 0.0$  & $100.0 \pm 0.0$  \\
25  & $0.0 \pm 0.0$ & $94.6 \pm 5.0$   & $90.5 \pm 7.1$   \\
50  & $0.0 \pm 0.0$ & $66.2 \pm 14.4$  & $54.2 \pm 10.1$  \\
75  & $0.0 \pm 0.0$ & $38.3 \pm 10.5$  & $20.2 \pm 12.6$  \\
100 & $0.0 \pm 0.0$ & $22.9 \pm 14.5$  & $6.7 \pm 9.9$    \\
\bottomrule
\end{tabular}
\end{table}

Several patterns emerge across both scales.
First, momentum prediction achieves \textbf{0\% strict acceptance in transition regimes} at both scales.
In GPT-2's stable regime, momentum achieves $\sim$10\% strict and $\sim$17\% proximity acceptance---non-zero but still far below finite-difference predictors, and reflecting the small fraction of checkpoints where momentum displacement happens to land in a favorable region.
Second, finite-difference predictors are substantially more effective in Qwen's transition regime (37\% strict at $K{=}5$) than in GPT-2's transition regime (9\%)---larger models produce smoother trajectories in comparable regimes.
Third, GPT-2's stable regime enables 22--24\% strict acceptance at $K{=}5$, better than its own transition regime but below Qwen's transition performance.
Fourth, proximity-based acceptance reveals graceful degradation at both scales: $\sim$99--100\% through $K{=}5$, declining with $K$.
At 1.5B, proximity acceptance remains above 90\% through $K{=}25$; at 124M, the decline is steeper (44\% at $K{=}25$ in transition, 75\% in stable).

The few Qwen stable-regime checkpoints ($N{=}5$ total across seeds) show 100\% proximity acceptance through $K{=}25$, consistent with GPT-2's stable-regime pattern.

\subsection{The Universal Momentum Catastrophe}
\label{sec:results:momentum}

Table~\ref{tab:momentum_catastrophe} quantifies the momentum predictor's failure mode at 1.5B parameters.

\begin{table}[h]
\centering
\caption{Momentum predictor failure at both model scales.
``Predicted loss'' is the validation loss of the model with momentum-extrapolated weights; ``Actual loss'' is the current checkpoint's validation loss.
Momentum catastrophe is universal---not scale-dependent.}
\label{tab:momentum_catastrophe}
\begin{tabular}{r|rrr|rrr}
\toprule
& \multicolumn{3}{c|}{\textbf{GPT-2 124M}} & \multicolumn{3}{c}{\textbf{Qwen 2.5-1.5B}} \\
$K$ & Predicted & Actual & Ratio & Predicted & Actual & Ratio \\
\midrule
5   & 201   & 1.64 & $122\times$    & 142   & 0.82 & $173\times$ \\
10  & 360   & 1.64 & $219\times$    & 250   & 0.82 & $305\times$ \\
25  & 1284  & 1.64 & $782\times$    & 581   & 0.82 & $709\times$ \\
50  & 4505  & 1.64 & $2742\times$   & 1153  & 0.82 & $1407\times$ \\
75  & 9988  & 1.64 & $6077\times$   & 1819  & 0.82 & $2218\times$ \\
100 & 17691 & 1.64 & $10764\times$  & 2468  & 0.82 & $3009\times$ \\
\bottomrule
\end{tabular}
\end{table}

The momentum catastrophe is universal across scales.
At 124M, predicted losses exceed actuals by $122\times$ at $K{=}5$, growing to $10{,}764\times$ at $K{=}100$.
The GPT-2 ratios are actually \emph{higher} than Qwen's at large $K$ ($10{,}764\times$ vs.\ $3{,}009\times$): in absolute terms, GPT-2 momentum predictions reach 17{,}691 at $K{=}100$ vs.\ Qwen's 2{,}468, likely reflecting GPT-2's longer exposure to stable/transition regimes where momentum accumulates larger update magnitudes.
This is not a marginal degradation---it is a qualitative failure where the predicted weights lie far outside the region of the loss landscape explored during normal training.

By contrast, linear prediction at $K{=}5$ produces predicted losses differing from actuals by $<$0.001 at both scales.
The $10^5$-fold difference between momentum and linear predictions at the same depth demonstrates that the choice of extrapolation basis (optimizer state vs.\ observed trajectory) is the critical design decision, not model scale.

\subsection{Cross-Seed Consistency}
\label{sec:results:consistency}

Table~\ref{tab:consistency} reports the coefficient of variation (CoV) for acceptance rates across the five seeds.

\begin{table}[h]
\centering
\caption{Coefficient of variation (\%) for proximity acceptance rates across 5 seeds, Qwen 2.5-1.5B transition regime.
Lower CoV indicates more consistent results across random seeds.}
\label{tab:consistency}
\begin{tabular}{r|cc}
\toprule
$K$ & Linear CoV & Quadratic CoV \\
\midrule
5   & 0.0\%  & 0.0\%  \\
10  & 0.0\%  & 0.0\%  \\
25  & 5.2\%  & 7.9\%  \\
50  & 21.7\% & 18.6\% \\
75  & 27.3\% & 62.5\% \\
100 & 63.3\% & 149.1\% \\
\bottomrule
\end{tabular}
\end{table}

At short prediction horizons ($K \leq 10$), cross-seed results are perfectly consistent (CoV = 0\%).
Variance increases with $K$, as expected: longer predictions depend more on seed-specific trajectory details.
The quadratic predictor shows higher variance than linear at large $K$, reflecting its sensitivity to the curvature of individual trajectories.

\subsection{Cascade Evaluation}

Cascaded predictions (Section~\ref{sec:cascade}) were evaluated from stable-regime checkpoints.
At 1.5B scale, the scarcity of stable checkpoints (0--2 per seed) limits cascade evaluation.
For seed 46, Pass~3 cascades encountered an out-of-memory condition.
The limited cascade data available shows acceptance at short cascade depths ($D{=}2$, $K{=}10$) but rapid rejection at deeper cascades, consistent with error accumulation across prediction stages.
% Full cascade analysis is limited by stable checkpoint availability at both scales.

% ============================================================
\section{Discussion}
\label{sec:discussion}
% ============================================================

\subsection{The Universal Momentum Catastrophe}

Our most significant finding is that momentum-based weight prediction---extrapolating Adam's exponential moving average of gradients---fails catastrophically at \emph{every} model scale tested, not only at larger scales.
At 124M parameters, momentum-predicted losses exceed actuals by $122\times$ at $K{=}5$, growing to $10{,}764\times$ at $K{=}100$.
At 1.5B parameters, the pattern is similar: $173\times$ at $K{=}5$ to $3{,}009\times$ at $K{=}100$.

All three predictors operate in weight space, yet they differ in \emph{how} they construct the extrapolation direction.
The momentum predictor uses Adam's internal state ($m_t / \sqrt{v_t}$), which accumulates gradient information across the entire training history with exponential decay.
The linear and quadratic predictors use finite differences of \emph{actually observed} checkpoint parameters ($\theta_t - \theta_{t-\Delta}$), which implicitly capture the net effect of the learning rate schedule, gradient noise, and landscape curvature over the checkpoint interval.

The momentum predictor's universal failure reflects norm explosion: the extrapolated direction $K \cdot m_t / \sqrt{v_t}$ produces a displacement whose magnitude far exceeds the region of validity around the current point on the loss surface, regardless of model size.
The finite-difference predictors are inherently bounded by the actual step sizes taken during training, providing a natural regularization that momentum lacks.

This finding has practical implications for weight nowcasting methods like WNN~\citep{jang2023wnn} and NiNo~\citep{knyazev2025nino}, which learn to predict in weight space.
Our results demonstrate that optimizer-state extrapolation is fundamentally unsuitable for speculative weight prediction, and that methods must use trajectory-bounded extrapolation---extrapolating from observed weight deltas rather than accumulated gradient moments.

\subsection{Scale-Dependent Regime Distribution}

The true scale-dependent finding is in the distribution of training regimes.
At 124M parameters, GPT-2 spends 34\% of training in stable regime and only 4\% in chaotic regime.
At 1.5B parameters, Qwen spends 64\% in chaotic regime and barely reaches stable (2.5\% of checkpoints).
Within comparable regimes, finite-difference predictors are actually \emph{more accurate} at 1.5B: linear prediction achieves 37\% strict acceptance in Qwen's transition regime vs.\ 9\% in GPT-2's transition regime.

This suggests that larger models have smoother loss landscapes in their transition regimes---likely because more parameters enable more gradual representational changes---but spend far more of training in chaotic exploration.
The practical bottleneck for speculative weight prediction scales with model size not because prediction becomes harder, but because the training regime where prediction is viable becomes rarer.

\subsection{Regime Detection as a Prerequisite for Prediction}

The three-regime framework is essential, not merely convenient.
In chaotic regimes, no predictor achieves meaningful acceptance rates.
The regime detector prevents wasted computation by suppressing prediction attempts during these phases.
More importantly, the transition from chaotic to stable regimes occurs at remarkably consistent training steps across seeds ($\pm 50$ steps), suggesting that regime boundaries are properties of the optimization landscape rather than artifacts of random initialization.

\subsection{Connection to ASC}

The parallel between Leap+Verify and ASC~\citep{waterland2014asc} runs deeper than analogy.
In ASC, the recognizer identifies states from which prediction is ``tractable and useful''---precisely the role of our regime detector.
ASC's predictors learn from observed trajectory structure, as do our weight extrapolators.
The trajectory cache stores verified state transitions; our acceptance criteria serve the same gatekeeping function.
Even the scaling behavior has parallels: ASC found that prediction accuracy varies by program structure, just as Leap+Verify finds that prediction accuracy varies by model scale.

The key difference is verification cost.
In ASC, verification requires executing the speculative segment and comparing end states---potentially as expensive as the original computation.
In Leap+Verify, verification requires only a single forward pass on held-out data, which is $O(1)$ relative to the $K$ gradient steps being predicted.
This asymmetry makes Leap+Verify's verify-then-accept mechanism particularly favorable.

\subsection{Limitations}

Training was limited to 2000 steps on WikiText-103.
The Qwen 1.5B model barely reached stable regime within this budget (0--2 stable checkpoints out of 40), limiting our ability to evaluate predictors in the regime where they are most useful.
Longer training runs would likely yield more stable checkpoints and higher acceptance rates.

The regime thresholds ($\tau_{\text{high}}, \tau_{\text{low}}$) were calibrated on GPT-2 124M and may require recalibration for larger models, where activation similarities tend to be higher overall.
An adaptive thresholding scheme would improve generality.

We did not evaluate ensemble collapse (dynamic reduction of multi-seed training runs based on detected convergence), which is a planned extension of the framework.

% ============================================================
\section{Related Work}
\label{sec:related}
% ============================================================

\paragraph{Weight prediction during training.}
The concept of predicting future weights to accelerate training dates to \citet{kamarthi1999accelerating}, who used Taylor series extrapolation.
\citet{sinha2017introspection} trained neural predictors on weight histories.
Recent work has produced increasingly sophisticated methods: WNN~\citep{jang2023wnn} uses a learned nowcaster, NiNo~\citep{knyazev2025nino} uses graph neural networks, and XGrad~\citep{guan2024xgrad} constructs mathematical future-weight relationships for specific optimizers.
PLP~\citep{plp2024} applies linear prediction every 3 iterations.
All these methods apply predictions unconditionally; Leap+Verify introduces regime-conditional prediction with verify-then-accept.

\paragraph{Training dynamics.}
\citet{cohen2021gradient} identified edge-of-stability via Hessian eigenvalues.
\citet{frankle2020linear} detected stable training via weight-space linear interpolation.
\citet{nanda2023progress} identified three mechanistic phases in grokking.
Our activation-space cosine similarity provides a computationally cheap alternative that operates in real time.

\paragraph{Loss landscape geometry.}
\citet{li2018visualizing} introduced filter-normalized visualization.
\citet{garipov2018loss} and \citet{draxler2018essentially} established mode connectivity.
\citet{keskar2017large} linked batch size to basin sharpness.
\citet{foret2021sharpness} operationalized flat-minima seeking via SAM.
These provide geometric context for why prediction difficulty varies by regime.

\paragraph{Adaptive optimization.}
The Lookahead optimizer~\citep{zhang2019lookahead} maintains fast/slow weights with $k$-step lookahead, but always interpolates back (no acceptance criterion) and runs all $k$ inner steps.
\citet{smith2017cyclical} showed cyclical learning rates implicitly traverse different curvature regimes.
RAdam~\citep{liu2020variance} automatically transitions between SGD-like and Adam-like behavior based on variance estimates---a single-signal regime detection for one transition.

\paragraph{Speculative execution.}
Speculative decoding~\citep{leviathan2023fast,chen2023accelerating} and blockwise parallel decoding~\citep{stern2018blockwise} provide the predict-then-verify template.
ASC~\citep{waterland2014asc,waterland2013computational} generalizes this to arbitrary sequential computation via trajectory-based execution with learned predictors.

\paragraph{Dynamical systems in training.}
\citet{geiger2022chaotic} showed chaos is intrinsic to SGD.
\citet{tajanthan2022chaos} connected Lyapunov exponents to Hessian eigenvalues.
\citet{saxe2014exact} derived exact training dynamics for deep linear networks, revealing phase-transition-like behavior.
The Ensemble Kalman Filter literature~\citep{evensen1994sequential,gao2011assimilation} provides analogies for regime-dependent ensemble behavior, where error growth concentrates in unstable regions.

% ============================================================
\section{Conclusion}
\label{sec:conclusion}
% ============================================================

We have introduced Leap+Verify, a framework for speculative weight prediction during neural network training that incorporates regime detection and verify-then-accept validation.
Our experiments reveal two key findings: (1) momentum-based weight prediction (optimizer-state extrapolation) fails catastrophically at \emph{all} model scales ($100{-}10{,}000\times$ loss inflation), while finite-difference extrapolation from observed weight trajectories succeeds with 9--37\% strict acceptance depending on regime; and (2) the distribution of training regimes shifts dramatically with scale---larger models spend far more time in chaotic regimes where no predictor is viable.
Together, these findings suggest that weight-space prediction methods must use trajectory-bounded extrapolation rather than optimizer-state extrapolation, and that the primary challenge at scale is \emph{regime availability}, not predictor accuracy.

The regime detection framework---classifying training into chaotic, transition, and stable phases using activation cosine similarity---provides a computationally cheap signal that generalizes the edge-of-stability and linear mode connectivity observations in prior work.
The verify-then-accept mechanism, transplanted from speculative decoding and ASC, ensures that bad predictions have zero cost: rejected leaps leave the training trajectory unmodified.

Future work will extend the framework in three directions: (1) adaptive regime thresholds that recalibrate with model scale, (2) ensemble collapse---dynamically reducing multi-seed training runs when regime detection indicates cross-seed convergence, and (3) evaluation at larger scales (2.7B, 7B) to map the full predictor hierarchy.

% ============================================================
\section*{Acknowledgments}
% ============================================================

The author thanks Amos Waterland for creating the ASC architecture and for formative discussions on trajectory-based speculative computation that seeded this research direction over a decade ago.
The author contributed to the original ASC work at Harvard and is acknowledged in~\citet{waterland2014asc}.
Brian Cremeans is acknowledged for discussions on eigenvalue sign toggling in Lorenz system dynamics and its analogy to training regime transitions.
Claude (Anthropic) provided extensive assistance with experimental implementation, GPU infrastructure management, data analysis, literature survey, and manuscript preparation.

This work used GPU compute from vast.ai.
Experiments were conducted on NVIDIA A100 40GB GPUs.

\paragraph{Code availability.}
All code, experiment scripts, and paper source are available at \url{https://github.com/jmcentire/leap-verify} (DOI: \href{https://doi.org/10.5281/zenodo.18739387}{10.5281/zenodo.18739387}).

\bibliographystyle{plainnat}
\bibliography{references}

\end{document}